\documentclass[final,5p,times,twocolumn]{elsarticle}

\usepackage{lineno}
\usepackage{graphicx}
\usepackage{amssymb}
\usepackage{pdflscape}
\usepackage[english]{babel}
\usepackage[dvipsnames,svgnames,x11names]{xcolor}
\usepackage{booktabs,tabularx}
\usepackage{multirow}
\usepackage{hyperref}
\usepackage{url}            
\usepackage{booktabs}       
\usepackage{amsfonts}       
\usepackage{nicefrac}       
\usepackage{microtype}      
\usepackage{amsmath}
\usepackage{commath}
\usepackage[english]{babel}
\usepackage[ruled]{algorithm2e}
\usepackage{caption}
\usepackage{subcaption}
\usepackage{lipsum}
\usepackage{graphicx}
\usepackage{makecell}
\usepackage[printonlyused]{acronym}
\usepackage[section]{placeins}
\usepackage[compact]{titlesec} 

\SetEndCharOfAlgoLine{}
\usepackage[leftmargin=6em,rightmargin=12em,indentfirst=false]{quoting}
\usepackage{siunitx}

\usepackage[final]{changes}
\usepackage{float}

\usepackage{lineno}
\usepackage{tikz}
\usepackage[export]{adjustbox}

\usepackage{graphicx}
\usepackage[utf8]{inputenc}
\usepackage[export]{adjustbox}
\usepackage{wrapfig}
\usepackage{dcolumn}

\makeatletter
\newcolumntype{T}[3]{>{\textfont0=\the@{#1}{#2}{#3}}c<{\DC@end}}
\makeatother

\usepackage{pgfplots}
\pgfplotsset{width=10cm,compat=1.9}

\usepackage{array}

\newcolumntype{L}[1]{>{\raggedright\let\newline\\\arraybackslash\hspace{0pt}}m{#1}}
\newcolumntype{C}[1]{>{\centering\let\newline\\\arraybackslash\hspace{0pt}}m{#1}}
\newcolumntype{R}[1]{>{\raggedleft\let\newline\\\arraybackslash\hspace{0pt}}m{#1}}

\usepackage{subfiles}

\usepackage{todonotes}

\RestyleAlgo{ruled}


\journal{Building and Environment}
\begin{document}
	
\begin{frontmatter}

\title{Cohort comfort models --- Using occupants' similarity to predict personal thermal preference with less data}

\author{Matias Quintana\,$^{1}$, Stefano Schiavon\,$^{2}$, Federico Tartarini\,$^{2}$, Joyce Kim\,$^{3}$,  Clayton Miller$^{1, *}$}

\address{$^{1}$College of Design and Engineering, National University of Singapore (NUS), Singapore}
\address{$^{2}$Center for the Built Environment, University of California Berkeley, U.S.}
\address{$^{3}$Department of Civil and Environmental Engineering, University of Waterloo, Canada}

\address{$^*$Corresponding Author: clayton@nus.edus.sg, +65 81602452}

\begin{abstract}
Cohort Comfort Models (CCM) are introduced as a technique for creating a personalized thermal prediction for a new building occupant without the need to collect large amounts of individual comfort-related data.
This approach leverages historical data collected from a sample population, who have some underlying preference similarity to the new occupant.
The method uses background information such as physical and demographic characteristics and one-time onboarding surveys (satisfaction with life scale, highly sensitive person scale, personality traits) from the new occupant, as well as physiological and environmental sensor measurements paired with a few thermal preference responses.
The framework was implemented using two personal comfort datasets containing longitudinal data from 55 people.
The datasets comprise more than 6,000 unique right-here-right-now thermal comfort surveys. 
The results show that a CCM that uses only the one-time onboarding survey information of an individual occupant has generally as good or better performance as compared to conventional general-purpose models, but uses no historical longitudinal data as compared to personalized models.
If up to ten historical personal preference data points are used, CCM increased the thermal preference prediction by 8\% on average and up to 36\% for half of the occupants in the first of the tested datasets.
In the second dataset, one-third of the occupants increased their thermal preference prediction by 5\% on average and up to 46\%.
CCM can be an important step toward the development of personalized thermal comfort models without the need to collect a large number of datapoints per person.
\end{abstract}

\end{frontmatter}


\section*{Nomenclature}
\renewcommand{\baselinestretch}{0.25}\normalsize
\renewcommand{\aclabelfont}[1]{\textsc{\acsfont{#1}}}
\begin{acronym}[longest]

\acro{pmv}[PMV]{Predicted Mean Vote}
\acro{pcm}[PCM]{Personal Comfort Model}
\acro{pcs}[PCS]{Personal Comfort System}
\acro{ccm}[CCM]{Cohort Comfort Model}
\acro{rhrn}[RHRH]{Right-Here-Right-Now}
\acro{hsps}[HSPS]{Highly Sensitive Person Scale}
\acro{swls}[SWLS]{Satisfaction With Life Scale}
\acro{rf}[RF]{Random Forest}
\acro{js}[JS]{Jensen-Shannon}
\acro{rbf}[RBF]{Radial Basis Function}
\acro{b5p}[B5P]{Big Five Personality}

\end{acronym}

\renewcommand{\baselinestretch}{1}\normalsize

\section{Introduction} 
In the built environment, indoor thermal comfort conditions may influence health~\cite{Ormandy2012, Pantavou2011}, office work performance~\cite{Hancock2007, Seppanen2006, Wyon2006, Zhang2019a}, learning performance~\cite{Mendell2005, Wargocki2019, Wargocki2017}, well-being~\cite{Lan2010, Altomonte2020}, and the overall satisfaction of occupants~\cite{DeDear2013, Zhang2019a, Parsons2014}.
An analysis done by Graham et al. (2021) based on approximately 90,000 occupant satisfaction survey responses found that roughly 40\% of them are dissatisfied with their thermal environment~\cite{20CBE-Graham2021}.

The Predicted Mean Vote (PMV)~\cite{Fanger1970} thermal comfort index is widely used in industry and research.
The PMV model aims to predict the average thermal sensation of a group of people sharing the same thermal environment.
Although Fanger~\cite{Fanger1970} was aware that people have different thermal sensations, he developed the PMV model since many buildings are not designed to provide personalized cooling and heating to individual occupants.
Hence, the PMV aims to find the optimal indoor conditions to ensure that most occupants are comfortable.
Another popular thermal comfort model included in the ASHRAE55 is the Adaptive model~\cite{RichardJ.deDear1998a}.
This model is based on the prevailing mean outside air temperature and only applies to naturally conditioned buildings.
Both models consider environmental and personal conditions affecting the occupant rather than specific temperature set-points as fixed comfort thresholds, but they average the individual occupants' responses.
A study on the ASHRAE Global Thermal Comfort Database II~\cite{FoldvaryLicina2018} provides evidence of the low accuracy of the PMV model, i.e., 33\%~\cite{Cheung2019} in predicting the thermal sensation of individual occupants.
This, combined with the proliferation of Internet-of-Things (IoT) sensors and wearable devices, has pushed researchers to look into alternative solutions to predict how occupants perceive their thermal environment.
Data-driven approaches leverage direct feedback from occupants paired with environmental and physiological data to develop personalized thermal comfort models~\cite{Kim2018_pcm-ml}.
These models can then be used to optimize the operation of Heating, Ventilation, and Air-conditioning (HVAC) systems or to quantify the thermal environment's quality in buildings.
With Personal Comfort Model~\cite{Kim2018_pcm-ml} researchers can use environmental, physiological, and behavioral data to predict thermal sensation or preference.
To investigate relationships between the variables mentioned above, field experiments offer a more realistic context of a person's experiences compared to most climate chamber studies~\cite{Zhang2010}.
Collecting field data poses several challenges since people must complete surveys while performing day-to-day tasks.
This may disrupt their activities, and accurate monitoring and logging of environmental variables with a high spatial and temporal resolution are expensive and nontrivial.
Moreover, even when there is enough available data to develop a PCM for all occupants, it is still problematic from the control strategy perspective to ensure that all occupants will find the thermal environment comfortable~\cite{Shin2017}.
Another common approach is to develop a general-purpose model to make the most of the available data.
This model is trained on the whole population of occupants so that the available data for training is maximized at the expense of personalization.
In open shared spaces where the HVAC system cannot be designed to tailor to individual needs, one option is to control environmental parameters to maximize the number of comfortable occupants and, at the same time, provide Personal Comfort Systems (PCS).
Each occupant can use the PCS (e.g., heated and cooled chairs, desk fans, and foot warmers) to adjust the environment based on their needs and preferences.

In this paper, we seek to investigate whether we can leverage historical information gathered from building occupants to predict the thermal comfort preferences of a new person from which little individual data is available.
By individual data, we refer to longitudinal subjective responses to Right-Here-Right-Now (RHRN) surveys about thermal preference.
In particular, our research hypothesis is that the thermal comfort preferences of building occupants can be estimated by analyzing their psychological and behavioral traits and comparing them to other occupants from which subjective feedback was previously obtained.
This way, we leverage the personalization aspect of PCM with less data.

In previous studies, the principal value of subjective responses alone was used to group occupants (i.e., the proportion of subjective feedback that is \textit{prefer cooler}, \textit{no change}, or \textit{prefer warmer})~\cite{Jayathissa2020} whereas in~\cite{Lee2017a} a \textit{comfort profile} was first determined based on some historical data. 
The new occupants were compared against the available group profiles.
In both works, once similarities were found between occupants, the group's data was used for the prediction model for new occupants. 
Environmental measurements (e.g., air temperature and relative humidity) coupled with physiological measurements (e.g., height and weight) were used as features in the model~\cite{Luo2020}.
However, in both approaches, some data is still required from the new occupant to correctly \textit{assign} it to a respective group.

In this work we introduce \acfp{ccm}.
This framework uses similarities between occupants to predict how new occupants with little or no historical data would perceive their environment.
More details are explained in Section~\ref{ccm}.
With this approach we tried to answer the following questions:

\begin{enumerate}
    \item Do \ac{ccm} increase the prediction performance for new occupants compared to a data-driven general-purpose model, i.e., all data are merged to develop a single comfort model?
    \item Can we pinpoint which occupants benefited from using \ac{ccm}, and if so, which features mainly contributed to improving the prediction accuracy?
\end{enumerate}

\section{Cohort Comfort Models}\label{ccm}
\subsection{Definition}

\begin{figure*}[htb!]
    \centering
    \begin{subfigure}[t]{\textwidth}
        \centering
        \includegraphics[width=0.5\linewidth]{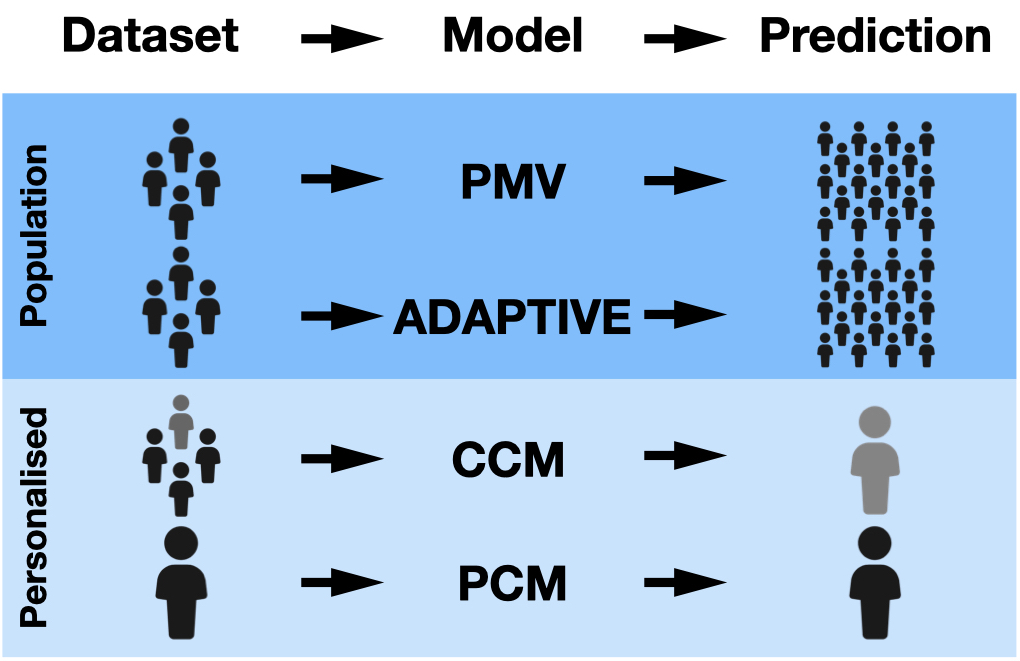}
        \caption{
            Diagram of thermal comfort modeling alternatives.
            The darker blue region highlights thermal comfort models at the population level (i.e., \acf{pmv} and adaptive).
            The light blue area comprises personalized thermal comfort models trained on the person itself (\acf{pcm}) or on a group of similar people (\acf{ccm}).
        }
        \label{fig:tc_overview}
    \end{subfigure}
    ~
    \begin{subfigure}[t]{\textwidth}
        \centering
        \includegraphics[width=.7\linewidth]{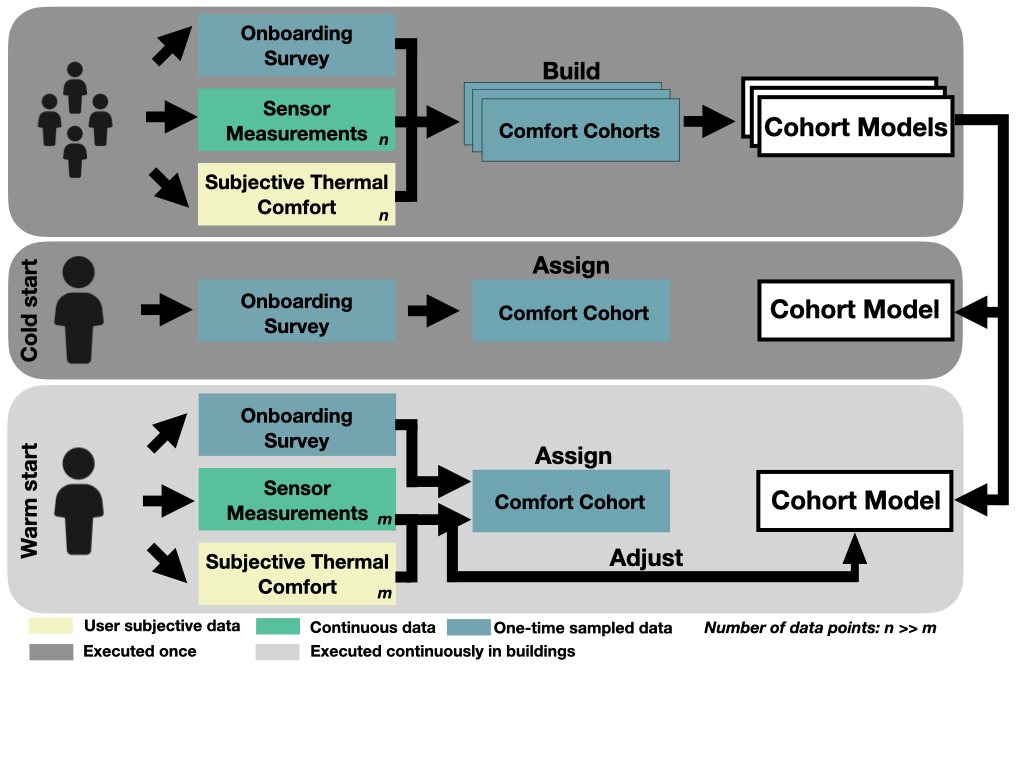}
        \caption{
        Cohort comfort modeling framework, built on \acfp{pcm}, leverages historical data collected from a sample population, who have some underlying preference similarity, to predict thermal preference responses of new occupants. 
        Firstly the comfort cohorts are created as shown in the top segment following a one-time data collection effort.
        Comfort cohorts are determined using data from the onboarding surveys, sensor measurements, and subjective thermal comfort data.
        The middle dark gray segment shows a new occupant from which only the onboarding survey is acquired and is used to assign the occupant to one of the previously created cohorts; this is also known as ``cold start''. 
        Finally, the bottom segment shows another new occupant from which $m$ data points are collected in order to be assigned to an existing cohort with the possibility of using these $m$ data points to fine-tune the existing cohort model; this is also known as ``warm start'' when $n >> m$. 
        For both cold and warm start, the model corresponding to the assigned cohort is used to predict the occupant's thermal comfort with the addition that the latter can use the already collected information from the occupant to fine-tune the cohort model.
        }
        \label{fig:ccm_overview}
    \end{subfigure}
    \caption{Comparison of current thermal comfort modeling alternatives (\ref{fig:tc_overview}) and overview of our work \textit{\acf{ccm}} (\ref{fig:ccm_overview}).}
    \label{fig:tc_ccm_overview}
\end{figure*}

\acfp{ccm} is a new framework to predict the thermal preference responses of a new person.
\ac{ccm} builds on \acfp{pcm} and leverages historical data collected from a sample population, who have some underlying preference similarity with the new person.
By doing this, \ac{ccm} leverage the power of both aggregate models and \acp{pcm}.
The key characteristic of a \acp{ccm} is that it is trained on a group of occupants who share features and subjective preferences with the new person we are trying to predict.
The comparison of data-driven thermal comfort modeling alternatives against more established ones like \ac{pmv} and the adaptive model is shown in Figure~\ref{fig:tc_overview}. 
An overview of the proposed framework is shown in Figure~\ref{fig:ccm_overview}.

One of the main contributions of this work is to determine ways in which characteristics of occupants, relevant to thermal comfort, can be compared. 
Firstly, a longitudinal data collection experiment of various buildings occupants is required, since a high number of participants allows the framework to find both occupants with similar and different characteristics. 
As for the required number of labeled data points per occupant ($n$ in Figure~\ref{fig:ccm_overview}), previous relevant works suggests that 50~\cite{Li2017}, 60~\cite{Kim2018_pcm-ml}, 90~\cite{Daum2011}, and 200~\cite{Liu2019} data points per occupant are needed for acceptable accuracy and stable prediction. 
However, the current landscape on thermal comfort longitudinal field data collection experiments is very limited in terms of datasets~\cite{Andre2020} and only a handful of them contain a high number of data points ($n$) per participant. 
It should also be noted that the number of data points needed for each participant varies as a function of the range of environmental conditions a participant may experience.
While a high $n$ would be ideal, our proposed framework is not restricted to a threshold of available data.
Our framework then utilizes the different data streams from the dataset collected (top gray segment, Figure~\ref{fig:ccm_overview}): onboarding survey, sensor data, \ac{rhrn}.
\textit{Onboarding survey} refers to a one-time background survey that each occupant fills out only once before the data collection experiment starts.
This survey must gather personal data (e.g., sex, height, weight) and preferably should also include psychological tests, such as the Big Five personality test~\cite{Gosling2003}.
This test is also used for occupant similarity, in the healthcare monitoring domain, for well-being label prediction~\cite{Li2020a}.
\textit{Sensor Measurements} encompasses all environmental and physiological data collected when the participant completed the \ac{rhrn}.
\textit{Subjective Thermal Comfort} comprises the questions asked in the \ac{rhrn} survey. 
These three main data streams are used to define the \ac{ccm}.These \acp{ccm} are groups or clusters capable of reflecting preference patterns extrapolated from the relationship between \textit{Subjective Thermal Comfort} and \textit{Sensor Measurements}, which are longitudinal time-series data we will refer to as \textit{warm start}.
Then, \ac{ccm} obtained from \textit{Subjective Thermal Comfort} and\textit{onboarding Survey responses} will be refered to as \textit{cold start}.
The naming convention is the same used in the field of Recommender Systems.
A \textit{cold start} approach means starting off from scratch with no previous ``warm up''. 
In our case this means we start without asking the participant to complete any right-here-right-now surveys. 
This is a harder start.
On the other hand, a \textit{warm start} scenario assumes some initial initialization or ``warm up'' has started, thus making the subsequent efforts easier.
In our case, this means we started with historical longitudinal data.

Once the cohorts are determined and existing occupants are allocated to one, new occupants can also be assigned to a cohort.
The cohort assignment is coupled with how the cohort was created and will vary depending on the data availability for the new occupant.
When a new occupant has not been through a longitudinal data collection phase, i.e., only onboarding survey data is available, they can be assigned to a cohort using a ``cold start'' assignment procedure (middle dark gray section in Figure~\ref{fig:ccm_overview}).
On the other hand, when a new occupant has been through a longitudinal data collection phase and a subset of \ac{rhrn} data points is available (bottom light gray section in \ref{fig:ccm_overview}), the assignment to a cohort can be done using a ``warm start'' approach. 
It should be noted, that in the latter case, the new occupant can still be assigned to its respective cohort without leveraging the collected labeled data points using a ``cold start'' assignment procedure with the onboarding survey responses. 
In addition, the availability of the \ac{rhrn} data for this occupant makes it possible to fine-tune the respective cohort model for the new occupant if required.
The details on \textit{cold start} and \textit{warm start} cohort creation and assignations are further explained in Section~\ref{cohort_creation_framework}.

\subsection{Current Related Work}
\subsubsection{Data-driven thermal comfort prediction}
Data-driven thermal comfort models rely on a handful of the measured features and often outperform the \ac{pmv} and adaptive comfort models~\cite{kim2018personal}, specifically when trying to predict the thermal preferences of individuals~\cite{Kim2018_pcm-ml}.
Recent literature employs machine learning models in order to contextualize environmental data~\cite{sgp-Cheung2017, Barrios2017, Gao2013, spot} as well as thermoregulation from human skin through video~\cite{jazizadeh2016can}, and physiological data~\cite{Liu2019, mansourifard2013online, zhang2014strategy}.
In addition, some work has also analyzed the transition time it takes for an occupant to change its thermal preferences in the same indoor environment~\cite{pinky2020}.
At the individual level,~\cite{Kim2018_pcm-ml} has shown that data-driven \ac{pcm} perform better than conventional models, such as the \ac{pmv}, on a cohort of 38 occupants. 
These \acp{pcm} are tailored to an occupant's specific needs and can be considered as one of the best approaches to achieving high-performing models for a particular individual.
As for the predicting variable, the 3-point Thermal Preference scale is a well-established scale used in thermal comfort research as the label or target variable for data-driven thermal comfort prediction modeling~\cite{Kim2018_pcm-ml}. 
Aggregated data-driven models, meaning models that use data from a large group of people to predict another person's thermal preference, face the challenge of multiple occupants having different thermal comfort preference from each other~\cite{Antoniadou2017, Brager2015, DeDear2013, VanHoof2010}.

\subsubsection{Cold start prediction}
The built environment research is moving towards using a small amount of data from new occupants in order to successfully predict their thermal comfort~\cite{Lee2017a}. 
This scenario is a well-known one when dealing with personalization, particularly in the field of \textit{Recommender Systems} and is known as ``cold start prediction''.
The cold start personalization problem refers to starting providing relevant information to the occupant, or a system, as fast and with as little effort as possible~\cite{Lika2014}. 
This is a common practice in smart agents (e.g., Cortana, Google Assistant, Siri) or services (e.g., Amazon, Netflix) where occupants can be clustered based on limited knowledge about them~\cite{Lam2008, Lika2014}. 
Alternatively, information about the occupants can be obtained by analyzing historical data and inferring other relevant features~\cite{Schein2002, Banovic2018}.
These two techniques can be used together to improve the model prediction performance.

In the context of health monitoring and well-being label (i.e., mood, health, stress) prediction, \cite{mHealth2020_published} provided a framework to quantify occupant similarity based on individual behavioral patterns and then cluster them in order to have groups from which group models can be used for new occupants.
\cite{Li2020a} used the Big Five personality trait survey as the one-time general preference survey to determine groups of occupants in a data-driven manner to then use group-level information for occupant personalization.
Both works rely on physiological, environmental, and behavioral data paired up with mood, health, and well-being labels from more than 200 different participants.
However, to the best of our knowledge, this is the first work that explores these techniques in the context of thermal comfort prediction.

\subsection{Cohort creation framework}\label{cohort_creation_framework}

\subsubsection{Cold start}\label{cohort_creation_survey}
Personal characteristics captured during the onboarding one-time survey, include categorical, ordinal, and scalar data that can directly be used to group occupants.
In addition, some standardized surveys allow for their questions scores to be aggregated, i.e., \acf{hsps}~\cite{Aron1997_sensory} and the \acf{swls}~\cite{Diener2006_swls}.

The simple approach to creating cohorts is to use individual one-time survey responses. 
Cohorts are created in a straightforward manner by dividing occupants based on a simple statistic, e.g., median, or creating as many cohorts as unique response values.
Multiple one-time survey responses can also be combined in a data-driven setup where each survey response is a feature and a clustering algorithm can automatically group occupants into cohorts based on the discovered patterns from the questions responses. 
We determined the optimal number of cohorts, $k$, by using the clustering metric known as Silhouette score~\cite{Rousseeuuw_SS} which ranges from 1 (best score) to -1 (worst score).
The appropriate $k$ is determined by the highest mean value of the Silhouette score.

Once the cohorts exist, new occupants can be directly assigned to one. 
If the cohorts were created in a data-driven manner, the trained clustering algorithm can then predict the corresponding cohort using the required occupant's one-time survey responses.
On the other hand, new occupants can be assigned to cohorts created based on specific values by simply matching its respective one-time survey response: e.g., if two cohorts are created based on sex, a new person is either assigned to the \texttt{Female} or \texttt{Male} cohort. 

\subsubsection{Warm start}\label{cohort_creation_longitudinal}
\begin{figure*}[htb!]
    \centering
    \begin{subfigure}[t]{0.49\textwidth}
        \centering
        \includegraphics[width=1\linewidth]{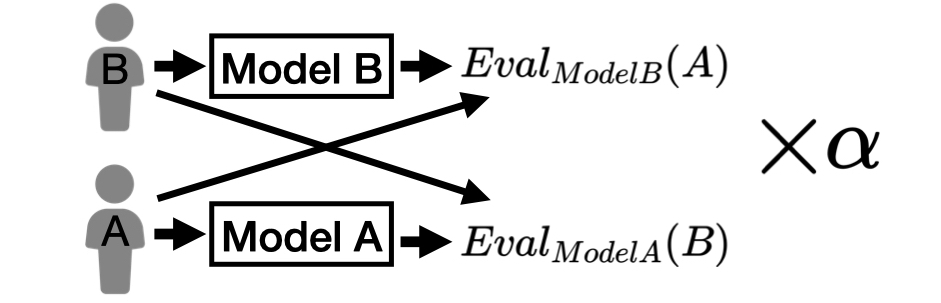}
        \caption{
            Cross-model performance evaluation calculated with F1-micro score, bounded in $[0,1]$, using each occupants' \ac{pcm} with another occupant' data. 
            The higher the F1-micro score the more similar the two occupants are.
        }
        \label{fig:cross_model}
    \end{subfigure}
    ~
    \begin{subfigure}[t]{0.49\textwidth}
        \centering
        \includegraphics[width=1\linewidth]{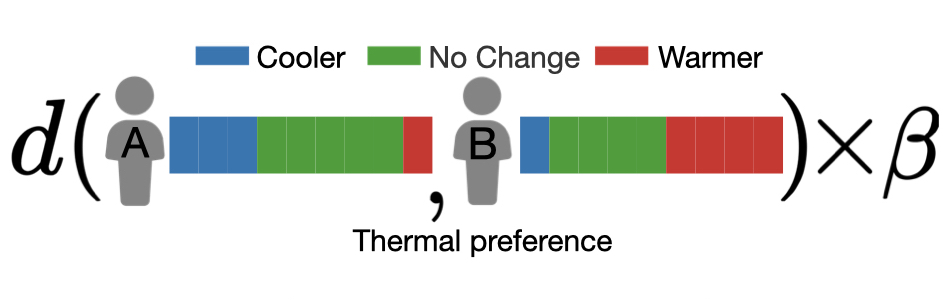}
        \caption{
            Responses distribution similarity based on Jensen-Shannon divergence, bounded in $[0,1]$. 
            After applying an \acf{rbf}, the higher the value the more similar the two occupants are.
        }
        \label{fig:dist_sim}
    \end{subfigure}
    \caption{
        \textbf{Warm start} cohort creation approach.
        \ref{fig:cross_model} calculates the cross-model performance and \ref{fig:dist_sim} calculates the distance between occupants' responses distribution.
        Each approach is weighted before their summation into an affinity matrix with $\alpha$ and $\beta$ respectively, where $\alpha + \beta = 1$.
        Spectral clustering is then run on the affinity matrix on a multiple number of cohorts (e.g., $k=[2, 10]$).
        The best $k$ is the one with the highest Silhouette Score.
        }
    \label{fig:sensor_clustering}
\end{figure*}

\textbf{Responses distribution similarity:}
Previous work has looked into grouping occupants solely based on the trend of their thermal comfort responses~\cite{Jayathissa2020}, for the case of thermal preference, occupants who predominantly have more \textit{prefer cooler} responses are grouped together, and so on. 
Figure~\ref{fig:dist_sim} illustrates the comparison of two occupants' thermal preference responses.
Unlike work done in~\cite{Jayathissa2020}, where grouping was done based on the majority response, the probability distribution of responses is first calculated for each occupant, on this example, the responses are thermal preference votes where the values are \textit{prefer warmer: -1, no change: 0, prefer cooler: 1}.
Then, the occupant's response distributions are compared using the \acf{js} divergence, bounded in $[0,1]$ and symmetric, which calculates the similarity between two probability distributions. 
A lower distance value between occupants means the occupants are more similar.

\textbf{Cross-model performance:}
We propose to group people into cohorts based on the relationship between their responses in the \ac{rhrn} survey and the different available environmental and physiological sensor data.
\cite{mHealth2020_published} suggest that these relationships can be captured by evaluating the model of one occupant, trained with its own data, using labeled sensor data from another participant, this is referred as cross-model performance evaluation.
In the context of thermal comfort, a \ac{pcm} is a good representation of an occupant's individual comfort requirements based on the data of its environment~\cite{kim2018personal}.
Hence, using \acp{pcm} for cross-model performance evaluation is an adequate similarity metric. 
If the \ac{pcm} of $occupant_A$ has a high performance on data from $occupant_B$, it suggests that both occupants have similarities in how they perceive their thermal environment.
Figure~\ref{fig:cross_model} shows an example of evaluating the cross-model performance of two occupants. 
The data used for the \ac{pcm} on each occupant comprise only sensor measurements since any personal characteristic data collected via a one-time preference survey would have a repeating value for each data point. 
For a consistent comparison, each \ac{pcm} must be trained on the respective occupant's data, and any hyper-parameter must be fine-tuned using the same evaluation metric. 
Based on the current literature on \acp{pcm} where the target variable comprises 3 ordinal values (i.e., multi-class), one of the most widely used models is \acp{rf}~\cite{Jayathissa2020, Kim2018_pcm-ml, Luo2020} and the evaluation metric is F1-micro score~\cite{Jayathissa2020, Luo2020} which can be seen as accuracy for the multi-class classification scenario. 
F1-micro scores range from 0 to 1, with 1 being a better prediction accuracy which in turn indicates two occupants are more similar.

\begin{figure*}[htb!]
    \centering
    \begin{subfigure}[t]{0.49\textwidth}
        \centering
        \includegraphics[width=1\linewidth]{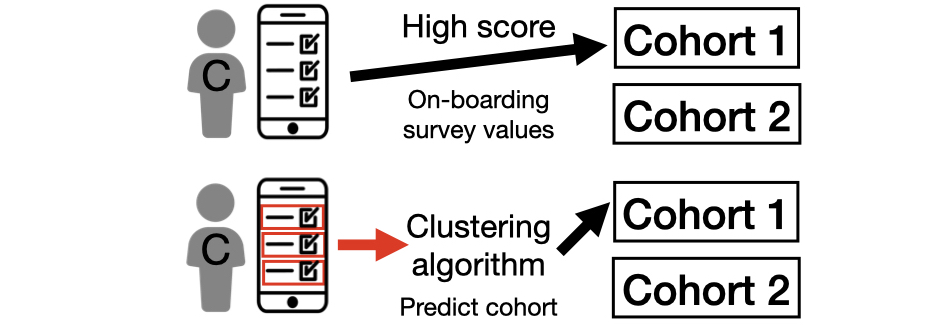}
        \caption{\textbf{Cold start}: \textbf{Top} The occupant is directly assigned to the cohort with the matching value of the onboarding survey used for cohort creation.
        \textbf{Bottom} The clustering algorithm used for cohort creation is used on the participant's onboarding survey responses to predict the corresponding cohort.
        }
        \label{fig:cohort_assignment_onboarding}
    \end{subfigure}
    \begin{subfigure}[t]{0.49\textwidth}
        \centering
        \includegraphics[width=1\linewidth]{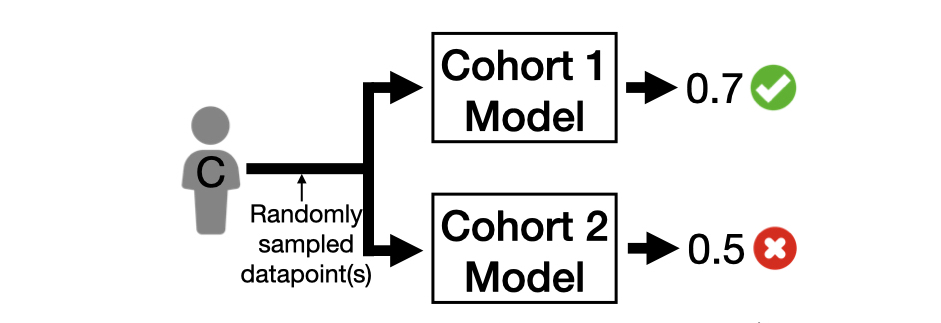}
        \caption{
        \textbf{Warm start}: Data points collected from the new participant are evaluated on each existing \textit{Cohort model}. 
        The participant is then assigned to the cohort on which it had the best-averaged performance based on F1-micro score. 
        }
        \label{fig:cohort_assignment_longitudinal}
    \end{subfigure}
    \caption{
        Participant assignment to the existing cohorts depending if the cohorts were cold start (\ref{fig:cohort_assignment_onboarding} or warm start (\ref{fig:cohort_assignment_longitudinal}). 
        Only 2 cohorts are shown for illustration purposes.
    }
    \label{fig:cohort_assignment}
\end{figure*}

Finally, both metrics of similarity between occupants, \textit{responses distribution similarity} and \textit{cross-model performance}, can be rearranged into what is called a similarity or \textit{affinity} matrix. 
An affinity matrix is a squared symmetric matrix where each element to compare, in this case occupants, is both the rows and columns indices. 
Each value of the matrix represents the similarity of its row and column; these values are bounded in $[0,1]$ where $1$ means identical elements.
Affinity matrices have shown to be a useful way of comparing occupants when the similarity calculations can be transformed into the $[0,1]$ range~\cite{mHealth2020_published}. 
Both similarity metrics already match the desired range but the responses distributions metric have an inverse interpretation where a value of $0$ means identical occupants. 
Thus, these values are normalized using the \acf{rbf} kernel (Equation \ref{eq:rbf}) such that now a value of 1 translates to identical occupants. 

\begin{equation}
    \label{eq:rbf}
    \text{RBF Kernel} \left(Z_{i}, c\right)=e^{\frac{-\left(z_{i, j}-c\right)^{2}}{2 \mu^{2}}}
\end{equation}

Where $Z_i$ is the similarity metric (\acf{js} divergence in this scenario), $\mu$ is the standard deviation of $Z_i$, $z_{i,j} \in Z_i$ is the JS divergence between occupants $i$ and $j$, and $c$ is the normalization center.
In order to consider both similarity metrics, both metrics can be added together with a weight coefficient $\alpha$ and $\beta$ (such that $\alpha + \beta = 1$), respectively, to indicate their respective contribution in the final affinity matrix. 
At this point, Spectral clustering~\cite{spectral_clustering} is applied to the affinity matrix and, similar to Section~\ref{cohort_creation_survey}, the adequate number of cohorts, $k$, is determined by the highest mean value of the Silhouette score. 
Figure~\ref{fig:sensor_clustering} illustrates both similarity metrics calculation, their rearrangement into an affinity matrix, and cohort calculation. 

New occupants can be assigned to a cohort only after they complete some \ac{rhrn} surveys.
However, as depicted in Figure~\ref{fig:ccm_overview}, collecting the same number of labeled datapoints, $n$, for each new occupant as it was done with the existing occupants can be impractical and cost prohibitive.
Hence, we propose to use a ``warm start'' approach that leverages a much smaller number of data points $m$ from the new participant, such that $m << n$. 
The $m$ datapoints from the new occupant are used to evaluate the performance of each cohort we previously created.
The performance values are then compared and the occupant is assigned to the cohort with the highest performance score overall. 
The metric used in this last step must be the same as the metric used to determine the initial cross-model performance evaluation, which for the case of a multi-class thermal comfort variable is the F1-micro score.
Figure~\ref{fig:cohort_assignment_longitudinal} illustrates this warm start cohort assignation.

\subsection{Evaluation}\label{evaluation}
Experiments and modeling parameters are detailed in Table~\ref{tab:exp_param}. 
We used a 0.8 train-test split, participant-wise, on each dataset.
Hence, 80\% randomly selected occupants and their data are used to build the cohorts following the approaches mentioned in the previous section.
The remaining 20\% of the occupants are used as test and assigned to a newly created cohort.
This entire process was repeated 100 times for each cohort approach on each dataset used.
Then, once a test occupant is assigned to a cohort, the cohort thermal preference model is evaluated on the new occupant's labeled data. 

\acfp{pcm} models were computed for cross-model performance and served as an upper-bound baseline. 
\acp{pcm} are trained with \ac{rf} as a the classification model following a 5-fold cross-validation.
We also used the entirety of the available data from the train participants to create an aggregate model where everyone belongs to one big cohort from which a \textit{general-purpose} model can be trained on. 
This model will serve as a lower bound baseline. 
All processes are repeated 100 times in order to minimize biases.

Additionally, the prediction results of \ac{pmv} at the personal level are included.
The Python package \textit{pythermalcomfort}~\cite{Tartarini2020a} is used to compute the \ac{pmv} as per the calculations methods in ISO 7730~\cite{iso20057730}, and the package \textit{Scikit-learn: Machine Learning in Python}~\cite{scikit-learn} is used for all the remaining data-driven calculations.
Similar to how it was done in~\cite{Kim2018_pcm-ml}, we used the longitudinal field data (i.e., air temperature, humidity, self-reported clothing insulation) and the static values (i.e., air velocity=0.1 m/s, metabolic rate=1.1met) for the PMV calculation unless the values are present in the dataset.
To compare the results on the same 3-point scale, the PMV is converted into thermal preference classes based on the following assumptions: $\left|\text{PMV}\right|\leq$ 1.5 is ``no change''; \ac{pmv} $>$ 1.5 is ``prefer cooler''; and \ac{pmv} $<$ -1.5 is ``prefer warmer''.
Nevertheless, we acknowledge that by making these many assumptions and simplifications on the values themselves, the accuracy of the PMV calculation will be low.
We share our code base for the reproducibility of our analysis on a public GitHub repository: \texttt{https://github.com/buds-lab/ccm}.

\subsection{Datasets}\label{datasets}
In order to assess the proposed framework, two publicly available longitudinal thermal comfort datasets were used to validate our methodology.
One dataset, named Dorn~\cite{TartariniDorn}, was collected in Singapore where 20 occupants were asked to complete approximately 1000 \ac{rhrn} surveys over a period of 6 months.
Participants were asked to wear a Fitbit smartwatch with the Cozie~\cite{cozie}\footnote{https://cozie.app/} application installed.
This application has been used to collect \ac{rhrn} surveys~\cite{Jayathissa2020} and is publicly available for both Android\footnote{https://github.com/cozie-app/cozie} and Apple\footnote{https://github.com/cozie-app/cozie-apple} platforms.
The second dataset we used was collected from 37 participants over a period of 12 weeks in an office building located in Redwood City, California, US~\cite{Kim2019PCS}.
Participants were asked to complete three \ac{rhrn} online surveys per day while using a \acf{pcs} in the form of an instrumented chair.
In this work we will be referring to the form dataset as the ``Dorn'' and the latter as the ``SMC'' dataset.
Both surveys comprised the following overlapping questions: thermal preferences (using a 3-point scale), clothing, and activity.
Environmental measurements (e.g., dry-bulb air temperature, relative humidity) were taken using data loggers installed in near proximity (i.e., within a 5-m radius) from where the participant completed the survey. 
Additionally,~\cite{TartariniDorn} used Fitbit Versa smartwatches to measure and log physiological data and iButtons for near-body temperature and humidity.
Table~\ref{tab:datasets_overview} shows a brief overview of both datasets, and Table~\ref{tab:sensors-accuracy} shows a summary of the sensors used and their accuracy.
More details regarding the recruitment criteria and methodology details can be found in the respective publications.

\begin{table}
    \centering
    \begin{tabular}{p{1cm}p{1.7cm}p{0.9cm}p{2cm}p{1.2cm}}
    \toprule
    Dataset & \shortstack{\# Occupants \\ (Sex)} & \shortstack{Age \\ range} & \shortstack{\#Responses per \\ participant} & Duration\\
    \midrule
    Dorn~\cite{TartariniDorn} & \shortstack{20\\(10 M, 10 F)} & \shortstack{20\\to 55} & \shortstack{872(min),\\1332(max)} & 6 months\\
    SMC~\cite{Kim2019PCS} & \shortstack{37\\(17 M, 20 F)} & \shortstack{25\\to 37} & \shortstack{33(min),\\218(max)} & 12 weeks\\
    \bottomrule
    \end{tabular}
    \caption{
        Overview of the two datasets used in this work.
    }
    \label{tab:datasets_overview}
\end{table}

Personal information about the participants (e.g., sex, height, weight) were obtained asking them to complete an on-boarding survey when joining the experiment.
Additionally, the Dorn dataset included three standardize surveys: \acf{hsps}~\cite{Aron1997_sensory}, \acf{swls}~\cite{Diener2006_swls}, and a short version of the \acf{b5p} test known as Ten-Item Personality Inventory~\cite{Gosling2003}. 
These surveys were used to investigate potential relationships between survey responses from occupants and their thermal preferences.

\subsection{Data pre-processing}\label{pre-processing}
We time-aligned sensor measurements with the \ac{rhrn} responses.
Thermal preference response provided by the occupants are used as the ground truth label for each respective dataset.
Feature selection scrutiny was done based on the latest efforts in data-driven personal thermal comfort prediction~\cite{Choi2012, Jayathissa2020, Kim2018_pcm-ml, Liu2019, Shan2020} and based on what the original dataset papers suggest~\cite{Kim2019PCS, TartariniDorn}.
A summary of the features used in this paper for model training is shown in Table~\ref{tab:features} for the present study, and Table~\ref{tab:features_dorn} and Table~\ref{tab:features_smc} for the Dorn~\cite{TartariniDorn} and SMC~\cite{Kim2019PCS} dataset respectively.

To have comparable results between occupants we decided to use a fixed number of responses per occupant on each dataset.
While this approach does not address the inherent class balancing issue in thermal preference datasets, i.e,. having a disproportionate number of data points for each thermal preference type~\cite{comfortGAN}, we decided to prioritize the quantity of available data points per participants.
The Dorn dataset contains a minimum of 872 responses per participant (Table~\ref{tab:datasets_overview}); however, after removing surveys completed outdoors, in non-transition periods and while participants were exercising, the minimum number of responses per occupant was 231. 
\cite{Kim2018_pcm-ml} found that when the dataset comes from an environment that all occupants shared during the experiment, the required number of responses for a stable thermal preference prediction is at least 60.
These results were obtained on the SMC~\cite{Kim2019PCS} dataset, hence, we opted to also use the first 60 responses from each occupant as it also covers most of its participants. 
This threshold of minimum responses resulted in 35 participants (18 females, 16 males).
Figure~\ref{fig:datasets_vote_dist} in the Appendix shows the distribution of the filtered thermal preference responses for each participant on both datasets.

\subsection{Cohort creation}
\label{cohort_creation_datasets}

Following the framework presented in Section~\ref{cohort_creation_framework}, we created two different sets of cohorts one for each datasets. 
Table~\ref{tab:cohort_datasets} shows the \textit{cold start} cohorts (upper row) and \textit{warm start} cohorts (lower row) that are constructed based on the data available in each dataset.
As mentioned in Section~\ref{cohort_creation_survey} the \textit{cold start} cohorts are created using only one-time survey responses. 
In the Dorn dataset, the onboarding surveys were also used to create other sets of cohorts. 
The three surveys (\ac{hsps}, \ac{swls}, and \ac{b5p}) were used together (\texttt{Surveys} in Table~\ref{tab:cohort_datasets}) where each survey score is treated as a feature and the number of cohorts was determined in a data-driven manner via Spectral clustering as detailed in Section~\ref{cohort_creation_survey}. 
Moreover, each survey was also used individually.
The \ac{hsps} and \ac{swls} surveys, named \texttt{Sensitive} and \texttt{Life Satisfaction} respectively in Table~\ref{tab:cohort_datasets}, have an aggregated numerical score which was used to create two cohorts based on the \textit{median} value of the occupants' scores. 
All participants were tested but only participants with extreme values on these survey responses, with a low and high aggregated score, are shown since their results are higher.
Occupants with an aggregated score between the 25\textsuperscript{th} and the 75\textsuperscript{th} interquartile range were filtered out, which meant only 50\% of the total occupants in the Dorn dataset were used.
Since the responses of the B5P cannot be aggregated into one final score, the number of cohorts based on this survey was determined in a data-driven manner using each personality score as a feature. 
This cohort approach is named \texttt{Personality} in Table~\ref{tab:cohort_datasets}.

Finally, \textit{warm start} cohorts are created following the two approaches mentioned in Section~\ref{cohort_creation_longitudinal}, using both the responses distribution similarity and cross-model performance combined, and cross-model performance alone (\textit{Dist-Cross} and \textit{Cross} respectively in Table~\ref{tab:cohort_datasets}). 
Table~\ref{tab:exp_param} list the different experimental and modeling parameters.
The best performing model, in terms of average expected F1-micro score, is found using a grid search on the list of hyperparameters.
In the Dorn dataset we determined that the optimal number of cohorts was 2 for all data-driven approaches.
On the other hand, a value of $k$ equal to 2 and 3 was chosen for the \texttt{Dist-Cross} and \texttt{Cross} approaches in the SMC dataset, respectively.
More details on different numbers of cohorts $k$ are shown in Figure~\ref{fig:dd-cohorts-datasets} in the Appendix.

In order to corroborate the usefulness of each cohort approach we also purposely assigned each occupant to the opposite cluster (if $k=2$ and were created by \texttt{cold start} approaches) or assigned the occupant to the worst performing cohort (\texttt{warm start} approaches).
This will serve as an ablation analysis for the cohort approach and, by extension, the data used for its creation and assignment. 
\vfill

\section{Results}\label{results}

\begin{table*}[hp]
    \begin{subtable}{\textwidth}
        \centering
        \begin{tabular}{cc}
            \toprule
            Feature & Source \\
            \midrule
            Air temperature & \multirow{2}{*}{Fixed sensor (indoor)}\\
            Relative humidity & \\
            \midrule
            Near-body temperature & \multirow{2}{*}{Wearable sensor} \\
            Heart rate & \\
            \midrule
            Clothing level & RHRN survey \\
            \midrule
            Sex & \multirow{3}{*}{One-time survey} \\
            Height & \\
            Weight & \\
            \bottomrule
        \end{tabular}
        \caption{
            Subset of features chosen for data-driven modeling in the Dorn dataset based on the original work~\cite{TartariniDorn}.
            Near-body (wrist level) temperature measurements are found to contain more information than skin temperature~\cite{Liu2019, Shan2020}.
        }
        \label{tab:features_dorn}
    \end{subtable}
    \vfill
    \begin{subtable}{\textwidth}
        \centering
        \begin{tabular}{cc}
            \toprule
            Feature & Source \\
            \midrule
            Dry-bulb air temperature & \multirow{4}{*}{Fixed sensor (indoor)}\\
            Operative temperature & \\
            Relative humidity & \\
            Slope in air temperature \\
            \midrule
            Control location & \multirow{6}{*}{PCS control behavior}\\
            Control intensity & \\
            Control frequency in the past x (1h, 4h, 1d, 1wk) & \\
            Occupancy status & \\
            Occupancy frequency in the past x (1h, 4h, 1d, 1wk) & \\
            \makecell{Ratio of control duration over occupancy \\duration in the past x (x=1 h, 4 h, 1 d, 1wk)} & \\
            \midrule
            Outdoor air temperature & \multirow{4}{*}{Outdoor Environment}\\
            Sky cover & \\
            Weighted mean monthly temperature & \\
            Precipitation \\
            \midrule
            Clothing level & \multirow{3}{*}{RHRN survey} \\
            Hour of the day & \\
            Day of the week \\
            \midrule
            Sex & \multirow{3}{*}{One-time survey} \\
            Height & \\
            Weight & \\
            \bottomrule
            \end{tabular}
            \caption{
                Subset of features recommended for data-driven modeling~\cite{Kim2018_pcm-ml} in the SMC dataset~\cite{Kim2019PCS}.
             }
            \label{tab:features_smc}
        \end{subtable}
    \caption{
        Features chosen for data-driven modeling: \ref{tab:features_dorn} Dorn~\cite{TartariniDorn}, \ref{tab:features_smc} SMC~\cite{Kim2019PCS} dataset. 
        Features obtained through the one-time survey are dropped for \acp{pcm} due to their constant value within each participant.
    }
    \label{tab:features}
\end{table*}

\begin{table*}
    \centering
    \begin{tabular}{ccc}
        \toprule
        Type & Cohorts approaches & Data used \\
        \midrule
        \multirow{5}{*}{Cold start} & Sex$\dagger$ & Self-reported value\\
        & Surveys & All onboarding surveys (\ac{hsps}, \ac{swls}, \ac{b5p})\\
        & Sensitive & \ac{hsps} scores\\
        & Life Satisfaction & \ac{swls} scores\\
        & Personality & \ac{b5p} scores\\
        & Agreeableness & Best performing trait from \ac{b5p}\\
        \midrule
        \multirow{2}{*}{Warm start} & \makecell{Responses \textbf{Dist}ribution  and \\ \textbf{Cross}-model performance (Dist-Cross)$\dagger$} & \makecell{Thermal preference responses\\ and occupants' \ac{pcm}}\\
        & \textbf{Cross}-model performance (Cross)$\dagger$ & Occupants' \ac{pcm}\\
        \bottomrule
    \end{tabular}
    \caption{
        Cohort approaches chosen. 
        The upper rows are cold start cohorts and the lower rows are warm start. 
        $\dagger$ were cohorts approaches used on both datasets. 
    }
    \label{tab:cohort_datasets}
\end{table*}

\subsection{Overall prediction performance}
\begin{figure*}[htb!]
    \centering
    \includegraphics[width=1\linewidth]{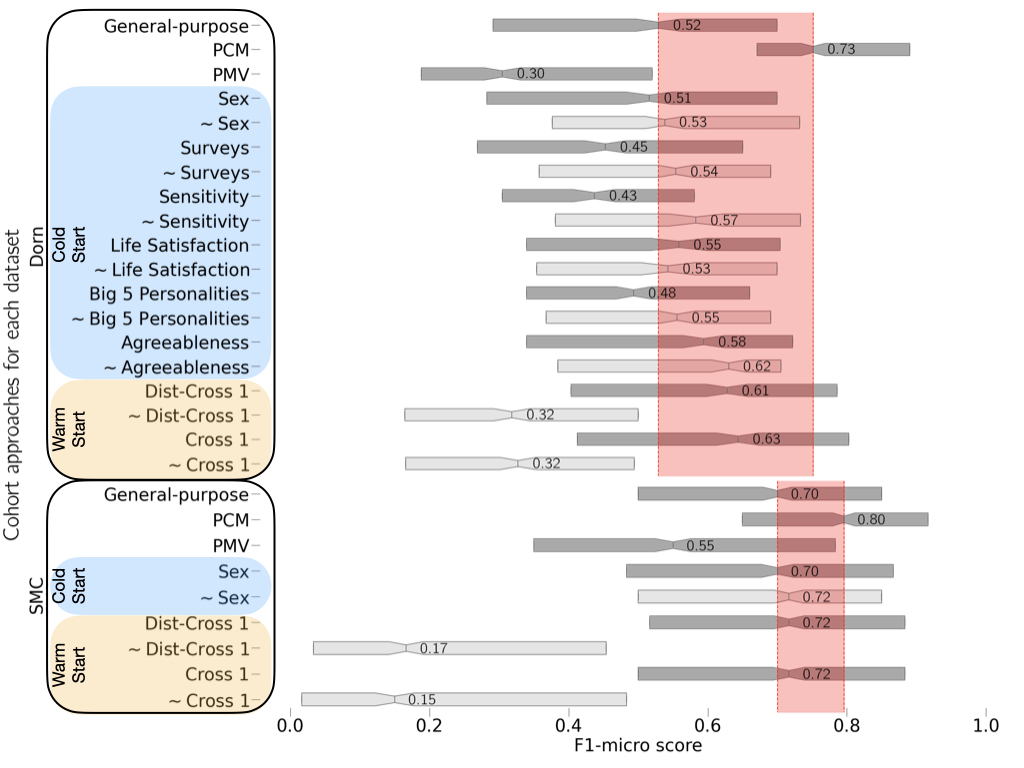}
    \label{fig:perf_comparison_dorn}
    \caption{
        Performance results in F1-micro score of 100 iterations for each cohort approach on each dataset Dorn and SMC (y-axis) and their baselines after correct assignment (dark gray) and incorrect assignment (light gray).
        Approaches with occupants incorrectly assigned to cohorts have the ``tilde'' symbol ($\sim$) before their name.
        \textit{Cold start} approaches are highlighted by a light-blue region and \textit{Warm start} approaches by a yellow region.
        The general-purpose model and PCM median value are highlighted with red dashed lines on each of their boxplot and filled with red.
        Additionally, PMV results, following the calculation in Section~\ref{evaluation} are shown.
        \textit{Warm start} approaches (\texttt{Dist-Cross} and \texttt{Cross}) have used only one randomly selected data-point.
        Dorn dataset: \texttt{General-purpose} boxplot contains 20 values; \texttt{Sensitive} and \texttt{Life Satisfaction} contain 200 values (2 test occupants at a time $\times$ 100 iterations) and all remaining box plots have 400 values (4 test occupants at a time $\times$ 100 iterations). 
        SMC dataset: \texttt{PCM} boxplot, contains 35 values and all remaining boxplots have 700 values (7 test uses at a time $\times$ 100 iterations). 
        The cohort approaches are taken from Table~\ref{tab:cohort_datasets}.
    }
    \label{fig:perf_comparison}
\end{figure*}

The performance results, in terms of F1-micro score, of all cohort approaches from Table~\ref{tab:cohort_datasets} in new occupants from their respective datasets are summarized in Figure~\ref{fig:perf_comparison}.
Results sets with a tilde ("$\sim$") as a prefix denote those approaches where test users were purposely assigned to an incorrect cluster as detailed in Section~\ref{cohort_creation_datasets}.

As mentioned in Section~\ref{cohort_creation_longitudinal}, \textit{Warm start} cohort approaches like \texttt{Dist-Cross} and \texttt{Cross} need that new participants complete a few \ac{rhrn} surveys prior to assigning them to a given cohort.
Different number of labeled data points, or \ac{rhrn}, were tested, i.e., 1, 3, 5, and 7, and all of them fall within the baselines.
Figure~\ref{fig:performance_appendix} shows the results for the Dorn dataset, similar results were obtained with the SMC dataset.
For the purpose of evaluating the minimum required number of labeled data points, all results in Figure~\ref{fig:perf_comparison} used one single, randomly selected, data point from each test occupant for assignment.
Compared to using more labeled data points, the performance of using a single data point still lead to improved results, on average 10\% compared to the general-purpose model.

\subsubsection{Under performing cohorts} 
Cohort approaches such as \texttt{Sex}, \texttt{Surveys}, \texttt{Sensitivity}, and \texttt{Big 5 Personalities} have a median performance below the \texttt{General-purpose} in the Dorn dataset (upper plot in Figure~\ref{fig:perf_comparison}).
Although ``sex'' has been found to have some influence in thermal preference, dividing occupants based on it does not provide significant changes, on average, on thermal preference prediction when compared to a \textit{general-purpose} model on both datasets.
This is supported by the similar performance of occupants being assigned to their same sex cohort and opposite sex cohorts (\texttt{Sex} and \texttt{$\sim$Sex} in Figure~\ref{fig:perf_comparison}).
The lower performance of two of the \texttt{Surveys} approach components (\texttt{Sensitivity} and \texttt{Big 5 Personalities}, when used individually, might be the reason for this approach's lower than \texttt{General-purpose} performance.
We hypothesize the \ac{hsps} survey, used for the \texttt{Sensitivity} was not able to capture enough information due to the geographical location and participants' background of the data collection experiment.
Participants from Singapore (Dorn dataset) may have provided a non-accurate response to this survey since they mostly experience a hot climate, compared to a full range of cold and hot climates, which could translate into the low performances of this cohort approach.
Each personality trait from the B5P survey was also tested individually and \texttt{Agreeableness} performed above the \texttt{General-purpose} value (Figure~\ref{fig:performance_appendix}).
This suggests that looking at all personality traits together (\texttt{Big 5 Personality}) is not as beneficial as looking into each personality trait individually.
Moreover, it is possible that the small sample size of participants on the Dorn dataset provided little variability on these survey responses or that the three onboarding surveys chosen, \ac{hsps}, \ac{swls}, and \ac{b5p}, are simply not relevant for thermal preference.
Overall, the efficacy and usability of these warm start approaches remains an open question.

\subsubsection{Above-baseline performing cohorts}
The \texttt{Life Satisfaction} is the only survey that by itself has an above \texttt{General-purpose} performance.
This indicates that treating each survey individually instead of combining them (\texttt{Surveys}) is a better approach for cohort prediction performance.
In fact, only the \ac{hsps} and B5P surveys lead to a cohort approach with favorable performances above the general-purpose models.
Since both of these cohort approaches are related to satisfaction and optimism, we hypothesize occupants within the resulting cohorts are less prone to preference changes due to small variations, meaning they have a wider acceptability range.

On the other hand, cohort approaches \texttt{Dist-Cross} and \texttt{Cross} achieved the highest median performance on both datasets (top and bottom plots in Figure~\ref{fig:perf_comparison} for the Dorn and SMC dataset, respectively). 
While one single labeled data point used for cohort assignment may be prone to bias, the consistent F1-micro scores after 100 iterations still positioned both cohort approaches as the top-performing ones, with a much clearer difference in the Dorn dataset than in the SMC dataset. 
Moreover, when each of this cohort approach is compared to their respective worst assigned cohort, the median performance is reduced by almost half for \texttt{Dist-Cross} (from 0.61 to 0.32) and \texttt{Cross} (from 0.63 to 0.32) in the Dorn dataset; and the performance is reduced 4 times for \texttt{Dist-Cross} (from 0.72 to 0.17) and \texttt{Cross} (from 0.72 to 0.15) in the SMC dataset. 
The median performance for \texttt{Life Satisfaction} is also reduced when occupants are incorrectly assigned (from 0.56 to 0.51).
The reduced number of occupants used for this approach, 10 instead of 20, might be one of the reasons why the difference in performance is less pronounced when compared to \texttt{Dist-Cross} and \texttt{Cross}
Nevertheless, the \texttt{Agreeableness} cohort approach exhibits different behavior.
Among all 20 occupants in the Dorn dataset, the lowest score obtained was 4 out of 7 which indicates the occupants rank highly on this personality trait. 
As previously mentioned, this personality trait might influence the acceptable thermal preference range of an occupant and, with only two cohorts created, both created cohorts could improve an occupant's thermal preference prediction.
These results suggest that when the cohort modeling framework is followed correctly, it leads to an overall increase in median F1-micro performance. 
This increase is shown to be higher with \textit{warms tart} approaches (\texttt{Dist-Cross} and \texttt{Cross}) than with \textit{cold start} approaches (\texttt{Life Satisfaction} and \texttt{Agreeableness}).

\subsection{Occupant-specific improvement}
\begin{figure*}[htb!]
    \centering
\includegraphics[width=1\linewidth]{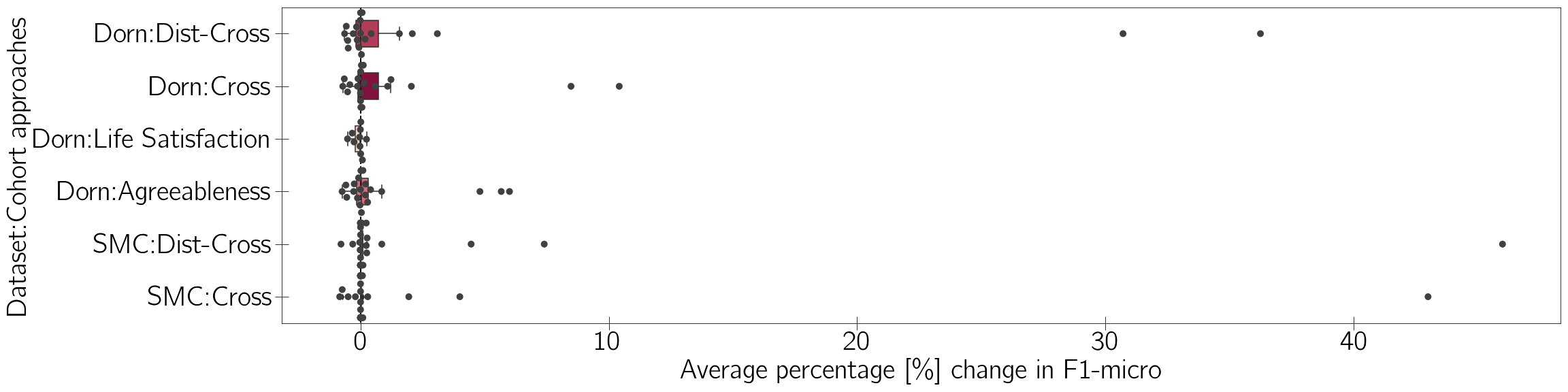}
    \caption{
        Scatter plot of occupant-specific percentage changes in F1-micro score from the general-purpose performance (\texttt{General-purpose}) to the respective cohort approach performance.
        Each dot represents the average value across all 100 iterations for one occupant.
        A positive value means the occupant benefited from the cohort approach and a negative one means they are worse-off.
        For the cohort approaches in their respective datasets (y-axis) 45\%, 60\%, 40\%, 55\%, 35\% and 32\% of the occupants saw an increase in their thermal preference prediction performance by relying on their peers' data, respectively, although the actual benefit is small.
    }
    \label{fig:occupant_change}
\end{figure*}

Figure~\ref{fig:occupant_change} shows a scatter plot of the average percentage changes in occupant-specific F1-micro score of the \textit{above-baseline performing} cohort approaches mentioned in the previous subsection for \textit{warm start} and \textit{cold start} types of cohorts on both datasets.

For each occupant, a positive average percentage change in F1-micro score is desirable since it indicates an increase in thermal preference prediction performance when using a cohort model instead of a general-purpose model.
Conversely, a negative value indicates the cohort model fails to provide useful personalization, and a generic model, which dismisses occupants' similarity entirely, performs better on average since it was trained with more data.
Overall, in the Dorn dataset around half of the population of occupants and one-third of the population of occupants in the SMC dataset saw a boost in prediction accuracy, in terms of F1-micro score, under the cohort modeling framework in their respective cohort approach.
For the remaining occupants, there is little change in performance.
In Figure~\ref{fig:perf_comparison} the cohort \texttt{Life Satisfaction} scored marginally better than the \texttt{General-purpose} (0.55 to 0.52) and the \texttt{Agreeableness} approached score a slightly higher median performance (0.58).
This is highlighted in Figure~\ref{fig:occupant_change} \texttt{Dorn:Life Satisfaction} where 40\% of occupants saw a performance boost of less than 1\%, whereas in \texttt{Dorn:Agreeableness} 55\% of occupants saw a performance boost of around 2\%.

\texttt{Dist-Cross} and \texttt{Cross} approaches show a performance boost of up to 36\% and 46\%, and an average of 5\% and 8\%, on the Dorn and SMC dataset respectively.
Overall, the number of occupants that benefited from these approaches are more than half (45\% and 60\%) and one-third (35\% and 32\%), respectively on each dataset.
The number of occupants that are better-off from using \texttt{Dist-Cross} or \texttt{Cross} are different on each dataset.
The \texttt{Dist-Cross} approach benefits more occupants than the \texttt{Cross} approach on the SMC dataset, where the data was collected in an office building across many multiple shared spaces on two different floors.
Under these circumstances, personal thermal preferences captured by the feedback distribution (the \texttt{Dist} component of \texttt{Dist-Cross}) contribute to more occupants being better-off by \texttt{Dist-Cross} compared to not using this information (\texttt{Cross} approach).
On the other hand, in the Dorn dataset all occupants provided their thermal preference response in their own work environments with no guarantees of similar conditions and exposures amongst all of the spaces. 
Occupants' feedback distribution (the \texttt{Dist} component of \texttt{Dist-Cross}) was not able to capture enough preference characteristics that could be quantified and compared among the occupants, resulting in more occupants being better-off with the \texttt{Cross} approach alone.
Nevertheless, it is also important to highlight that some occupants are actually worse-off regardless of the cohort approach used, and their average percentage performance is reduced. 
The entire distribution of percentage changes across all 100 iterations for these approaches and for each occupant can be found in Figure~\ref{fig:change_dist_dorn} and \ref{fig:change_dist_smc} for the Dorn and SMC dataset respectively.

Finally, metadata about the participants who are better-off and worse-off is summarized in Table \ref{tab:occupants_metadata}.
Overall, there are no major differences between the sex and mean and standard deviation of height and weight among the occupants in both groups, in most cohort approaches. 
The cohort approach of \texttt{Life Satisfaction} on the Dorn dataset and \texttt{Cross} on the SMC dataset show a slight majority of Female and Male better-off occupants, respectively.
Nevertheless, due to the small sample size of occupants on both datasets it is not possible to generalize these findings.

\section{Discussion}\label{discussion}
We showed that \ac{ccm} can improve the thermal preference prediction of a new occupant by using other occupants' data who are grouped based on preference similarity. 
We determined that: 
i) \textit{Warm start} cohort approaches outperform \textit{cold start} approaches at the cost of requiring labeled data points from each new occupant; 
ii) while around half and one-third of occupants, in each Dorn and SMC dataset respectively, benefited from being assigned to a cohort via warm start and using its cohort model, the remaining occupants saw a very little variation in their performance for both being better-off or worse-off; and
iii) the applicability of cohorts is dependent on the availability of data from the occupants as well as the number of occupants willing to provide it, as we found ourselves able to investigate more cohort approaches on the dataset with more sensed modalities and more one-time onboarding surveys, like in the Dorn dataset. 

\subsection{Warm vs cold start}
\textit{Warm start} cohorts are bottom-up approaches that focus on the granular data with a direct connection with thermal preference, i.e., relying on feedback distribution and cross-model performance, whereas \textit{cold start} cohorts are a top-bottom approach that relies on the set of questions used to showcase the separability of personal thermal preferences.
It can be argued that the latter approaches were expected to underperform since by their definition, information regarding the actual thermal preference of occupants, e.g., the labeled data points, was not used to create the cohorts. 
Nevertheless, recent literature in personalized healthcare monitoring \cite{mHealth2020_published, Li2020a} highlights the usability of these sources of information as primary ways to leverage other occupants' data.
We found that the \ac{swls} surveys, or \texttt{Life Satisfaction}, and the personality trait of \texttt{Agreeableness} from the \ac{b5p} survey lead to cohorts where thermal preferences are shared, leading to slightly improved thermal preference prediction when compared to general-purpose models.
One advantage of using this information is that allows a cold start from a new occupant.

However, the notion of ``no historical data'' is not a harsh threshold since depending on the amount of data available, some work still considers a cold start when very little historical data is used \cite{Banovic2018}, i.e., less than 10 data points \cite{mHealth2020_published}. 
While work done in \cite{Lee2017a} explores a Bayesian approach to cluster occupants based on historical measured data and then uses the cluster's thermal comfort models with as little as 8 data points which also falls under the cold start definition used in \cite{mHealth2020_published}, to the best of our knowledge our proposed framework is the first to not only look at measured data streams to find occupant similarity but includes the aforementioned mechanisms from other fields.
Our framework allows for a full cold start prediction and a warm start prediction with as little as one single historical data point.
Both types of cohort approaches have shown that while only some occupants benefited from this framework, the majority experience very little variability in performance, either better-off or worse-off.
Nevertheless, this also shows that our framework allows for prediction performance just as good as the general-purpose models but with at least half less training data, for the case of only two cohorts.


\subsection{From ``how much can you benefit'' to ``who can benefit''}
The \textit{warm start} cohorts at the occupant level produced an overall average increase of up to 8\% (\texttt{Dorn:Dist-Cross}) and 5\% (\texttt{SMC:Dist-Cross}) on the Dorn and SMC dataset respectively among the occupants who were indeed better-off.
These numbers changed to 4\% (\texttt{Dorn:Dist-Cross}) and 2\% (\texttt{SMC:Dist-Cross}) when all participants are considered in the calculation.
However, on both datasets there are occupants who could significantly take advantage of these approaches with performance increases as high as 45\%, but also occupants who saw a decrease in their prediction performance.
While this scenario is not uncommon when dealing with group-level occupant personalization \cite{Li2020a}, cold start approaches like our framework offers the possibility to identify these better-off and worse-off occupants even before the occupant starts occupying the building.

From a facility manager's perspective, being able to identify if an occupant will be worse-off, or better-off, on a specific thermal zone with another group of occupants can help prioritize certain occupants, avoid complaints, and quantify the need for PCS.
The trend of hot-desking is not well received among occupants \cite{Morrison2017} and can show a negative effect on work engagement, job satisfaction, and fatigue \cite{Hodzic2021}.
Thus, a more suitable approach would require engaging with specific occupants and adapt their needs.
On the other hand, although the results here are limited to two datasets, it is possible that some occupants are inherently more difficult to predict than others, and pinpointing back to these occupants alone is equally important.
Be that as it may, cold start approaches that rely on one-time surveys, such as \texttt{cold start} cohort approaches, raise data privacy concerns. 
Facility managers and building practitioners must take the necessary precautions when dealing with such data from its occupants.

\subsection{Limitations and future work}
The evaluation of our proposed framework based on one-time onboarding surveys and sensor measurements as features has some limitations.
First, the amount of occupants on each dataset is 10 times smaller than those used in related work in the healthcare domain \cite{Li2020a, mHealth2020_published} and in a related thermal comfort cluster prediction \cite{Lee2017a}.
Additionally, one of the two datasets used (Dorn dataset) lacks seasonal variability because it was collected in Singapore (tropical climate), whereas the other dataset (SMC) which included both seasons and a more diverse group of occupants, lacks the one-time surveys required by the framework. 
While multiple datasets can be used together as one much bigger dataset, like the ASHRAE Thermal Comfort Database II \cite{FoldvaryLicina2018}, the methodology and available measurements should overlap, which is not the case for the recent field data collection experiments.

Second, the models and features used were not chosen after an exhaustive feature and model selection pipeline. 
One could argue that features extracted from the physiological time series measurements from each occupant, matched with their thermal preference responses, could contain useful information which could in turn be used for comparison across occupants and subsequent cohort creation. 
While extracting those temporal attributes is out of the scope of the present work, the modularity of the framework and its data-agnostic proposition makes the inclusion of these newly learned or discovered features, or any other new modalities, plausible.

Third, in our evaluations, we considered that once an occupant is assigned to a cohort, it remains its membership for the duration of the evaluation experiment. 
It is likely that influenced by external factors, an occupant might be better-off in terms of thermal preference prediction if it is assigned to a different cohort in certain settings. 
Future work can build on this cohort framework and insert the respective adaptive mechanisms that allow occupants to change cohorts as time goes by. 
Particularly, when coupled with the energy repercussions of using cohort models to predict thermal preference, an entire feedback loop from the occupant to the system could enable this online incremental learning.

Aware of the context-dependent nature of our framework and its components, we do not claim generalizable results. 
Ideally, a dataset with a large enough group of people where all occupants undertake the same one-time onboarding set surveys, with overlapping sensor measurements, could provide more generalizable insights.
Also, if multiple datasets from different contexts are used, features that represent the location, climate, and building type could be considered as part of the modeling portion of the framework.
However, the presented framework is modular enough to easily adapt to new sets of features either as part of the cohort models themselves or as part of how the cohorts are created. 
We plan to investigate the results on more diverse datasets with occupants from different backgrounds and populations.


\section{Conclusion}
\label{conclusion}
\acfp{ccm} are a new approach to thermal comfort modeling that builds on \acfp{pcm}. 
Cohort models leverage the previously collected data from other occupants to predict an occupant's thermal preference based on similarities with other occupants.
The advantage of this method in two different datasets shows that the proposed framework for cohort creation and occupant assignment lead to half and one-third of the occupants in each dataset experiencing an average increase of 5\% and 8\%, respectively, in their thermal comfort prediction performance with as little as a single labeled data point required from the new occupant when compared to general-purpose data-driven models. 

Unlike related literature focused on occupant segmentation for better thermal preference prediction, our proposed framework offers the ability to identify the occupants who will be better-off from said cohort approach, and those who might not, based on the occupant's background information or, at most, a single labeled data point from the occupant at the building premises.
We provided the framework in a data and site-agnostic manner and described its different implementations depending on what information is available with the scalable potential to various data streams.
Cohort comfort models can benefit the building industry by improving the level of thermal comfort among occupants without the need to rely on individual customization which can be unfeasible and too expensive. 
Further global data collection experiments with multiple occupants are encouraged to investigate the generalization of the potential different cohorts and the potential discovery of thermal preference \textit{signatures} shared across different occupant populations.
Additionally, we encourage more research on control strategies that take advantage of the cohort-customization instead of individual-level catering.


\section*{CRediT author statement} 
\textbf{Matias Quintana:} Conceptualization, Methodology, Software, Validation, Formal analysis, Investigation, Visualization, Writing - Original Draft.
\textbf{Stefano Schiavon:} Conceptualization, Visualization, Supervision, Writing - Reviewing and Editing, Funding acquisition.
\textbf{Federico Tartarini:} Resources, Data Curation, Writing - Reviewing and Editing.
\textbf{Joyce Kim:} Resources, Data Curation, Writing - Reviewing and Editing.
\textbf{Clayton Miller:} Conceptualization, Resources, Visualization, Supervision, Project administration, Writing - Reviewing and Editing, Funding acquisition.


\section*{Acknowledgements}  
This research was funded by the Republic of Singapore's National Research Foundation through a grant to the Berkeley Education Alliance for Research in Singapore (BEARS) for the Singapore-Berkeley Building Efficiency and Sustainability in the Tropics (SinBerBEST2) Program and a Singapore Ministry of Education (MOE) Tier 1 Grant (Ecological Momentary Assessment (EMA) for Built Environment Research). 
BEARS has been established by the University of California, Berkeley as a center for intellectual excellence in research and education in Singapore.
The authors would like to acknowledge the team behind the data collection and processing, including Mario Frei, Yi Ting Teo, and Yun Xuan Chua.
\clearpage


\appendix{}
\section[pfx={Appendix},nonum]{}
\renewcommand\thefigure{\thesection.\arabic{figure}}
\renewcommand\thetable{\thesection.\arabic{table}}
\label{appendix}

\setcounter{figure}{0}
\setcounter{table}{0}

\begin{table*}
    \centering
    \begin{tabular}{cc}
        \toprule
        Parameter & Description or value\\
        \midrule
        Train-test ratio & 0.8, participant-wise\\
        \midrule
        Model & Random Forest \\
        \midrule
        Model Hyperparameters & \makecell{Number of trees: 100, 300, 500\\
                                        Split criterion: Gini index\\
                                        Tree depth: 1, $\dots$, 10\\
                                        Min. smaples for split: 2, 3, 4\\
                                        Min. samples on an edge : 1, 2, 3
                                        }\\
        \midrule
        Model training & \makecell{Hyperparameter grid search\\5-fold Cross-Validation} \\
        \midrule
        Model metric & F1-micro score \\
        \midrule
        Clustering & \makecell{Number of cohorts tested: $k \in [2, 10]$ \\Metric: Silhouette score}\\
        \midrule
        Iterations & Modeling pipeline is repeated 100 times\\
        \bottomrule
    \end{tabular}
    \caption{Experiments and modeling parameters}
    \label{tab:exp_param}
\end{table*}

\begin{figure*}[htb]
    \centering
    \begin{subfigure}{0.49\textwidth}
        \centering
        \includegraphics[width=1\linewidth]{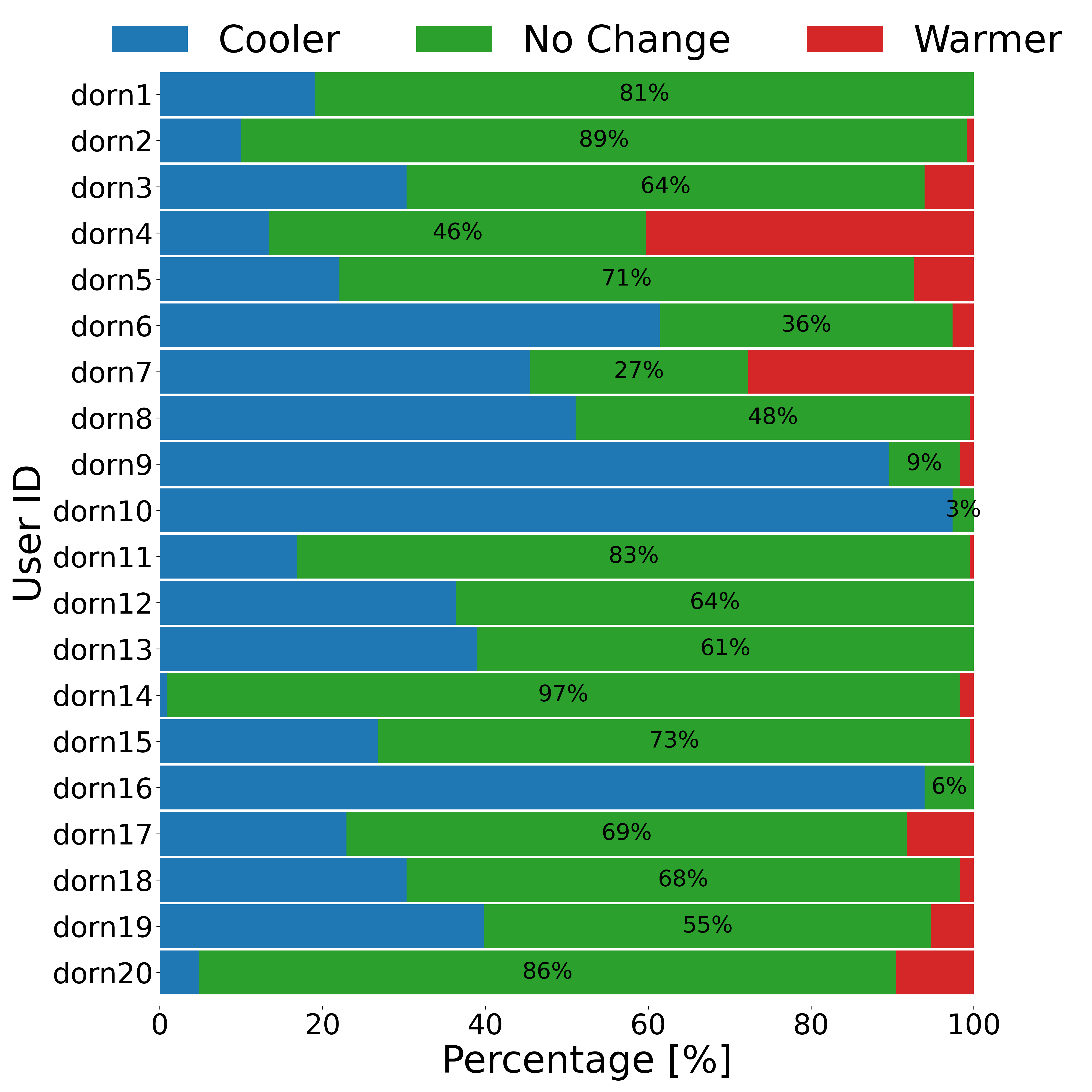}
        \caption{Dorn~\cite{TartariniDorn} participants and their first 231 responses per participants.}
        \label{fig:dorn_vote_dist}
    \end{subfigure}
    ~\hfill
    \begin{subfigure}{0.49\textwidth}
        \centering
        \includegraphics[width=1\linewidth]{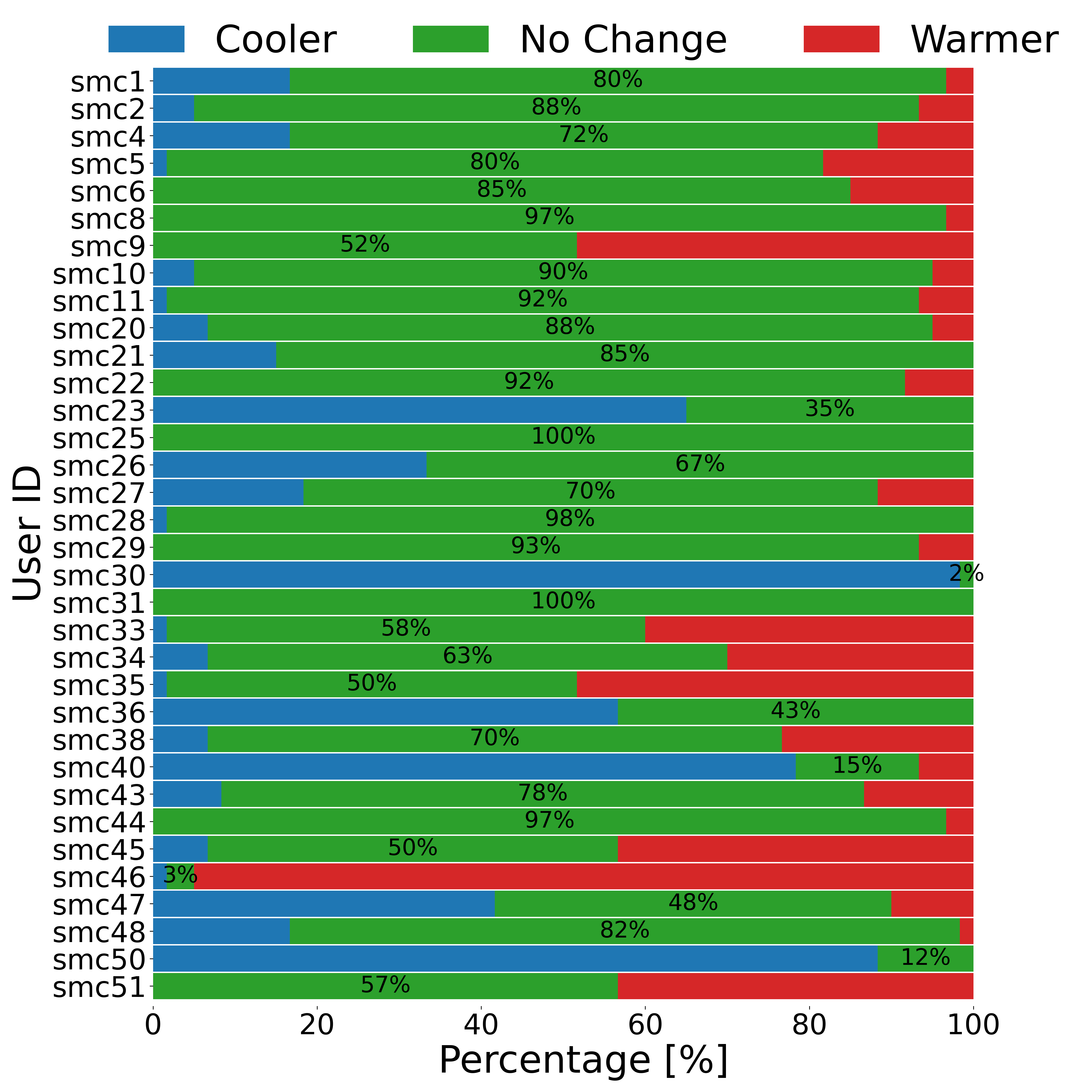}
        \caption{SMC~\cite{Kim2019PCS} participants and their first 60 responses per participants}
        \label{fig:smc_vote_dist}
    \end{subfigure}
    \caption{
        Thermal preference response distribution for all participants in both datasets. 
        \ref{fig:dorn_vote_dist} shows the Dorn~\cite{TartariniDorn} dataset and \ref{fig:smc_vote_dist} shows the SMC~\cite{Kim2019PCS} dataset.
    }
    \label{fig:datasets_vote_dist}
\end{figure*}

\begin{figure*}[htb]
    \centering
    \begin{subfigure}[t]{0.49\textwidth}
        \centering
        \includegraphics[width=1\linewidth]{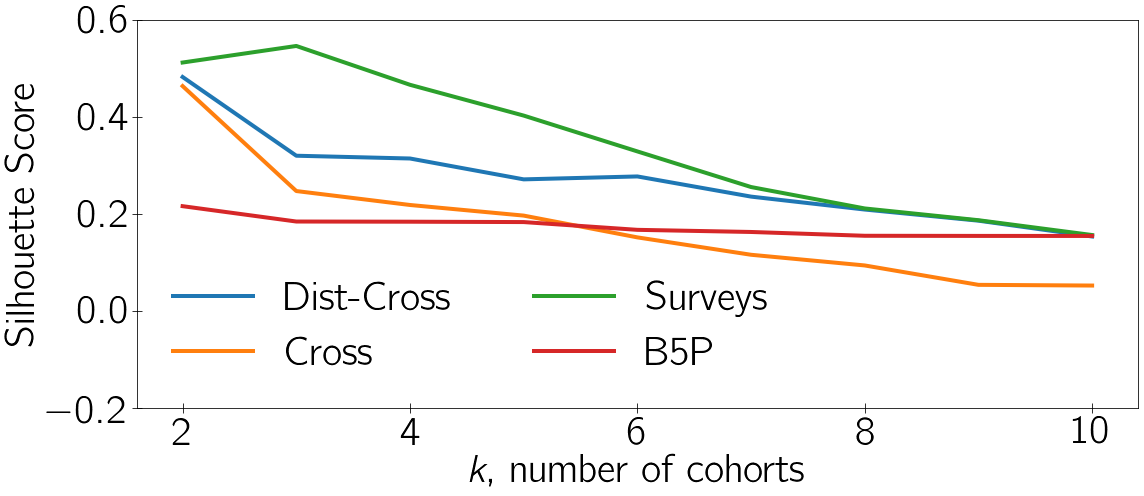}
        \caption{
            Average Silhouette scores for \textit{cold start} (\texttt{Surveys} and \texttt{B5P}) and \textit{warm start} (\texttt{Dist-Cross} and \texttt{Cross}) approaches on the Dorn dataset.
        }
        \label{fig:dd-cohorts-dorn}
    \end{subfigure}
    ~\hfill
    \begin{subfigure}[t]{0.49\textwidth}
        \centering
        \includegraphics[width=1\linewidth]{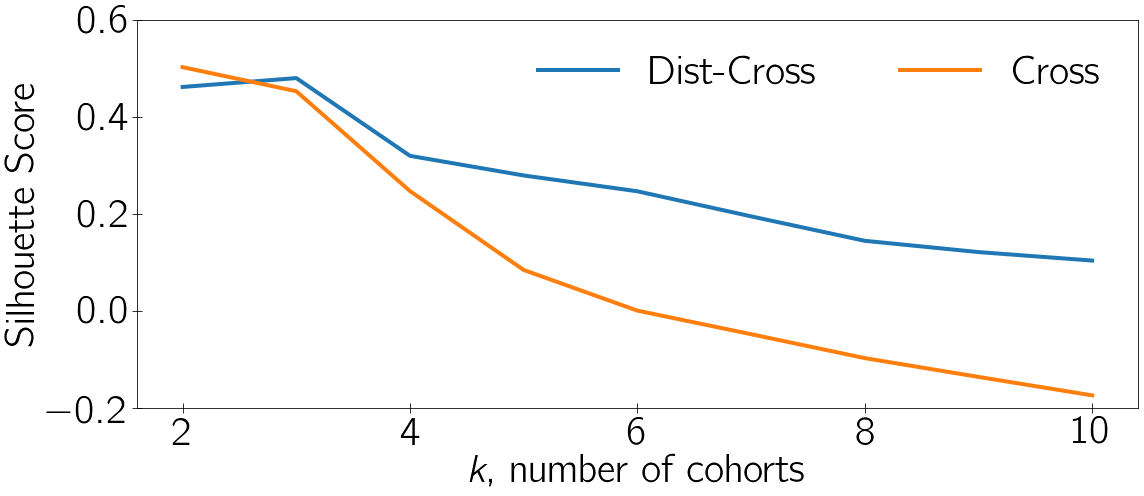}
        \caption{
            Average Silhouette scores for \textit{warm start} approaches on the SMC dataset.
        }
        \label{fig:dd-cohorts-smc}
    \end{subfigure}
    \caption{
        Number of cohorts determined based on the average Silhouette scores at different number of cohorts $k$ ($k \in [2, 10]$) on both datasets.
        The higher the Silhouette score the better.
        \ref{fig:dd-cohorts-dorn} and \ref{fig:dd-cohorts-smc} showcase the values for the Dorn and SMC datasets respectively.
    }
    \label{fig:dd-cohorts-datasets}
\end{figure*}

\begin{figure*}[htb]
    \centering
    \includegraphics[width=1\linewidth]{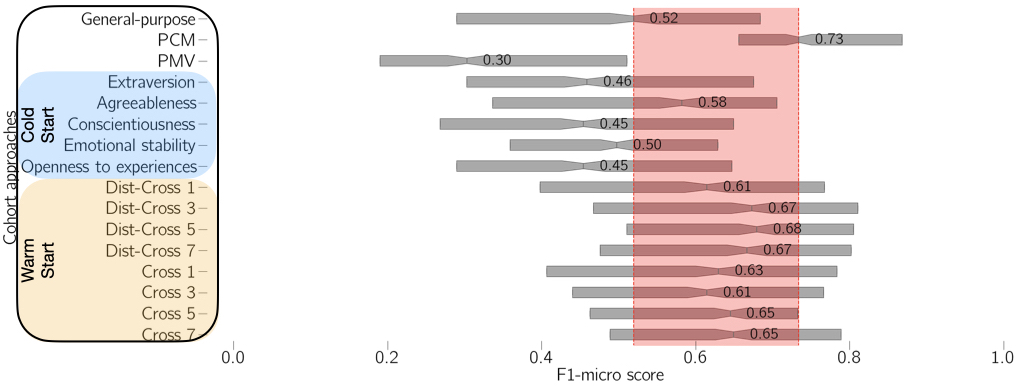}
    \caption{
        Performance results in F1-micro score of 100 iterations for each cohort approach on the Dorn dataset (y-axis).
        \textit{Cold start} approaches are highlighted by a light-blue region and \textit{Warm start} approaches by a yellow region.
        These results are complementary to Figure~\ref{fig:perf_comparison}.
        Each personality trait from the B5P (\texttt{Extraversion}, \texttt{Agreeableness}, \texttt{Conscientiousness}, \texttt{Emotional stability}, \texttt{Openness to experiences}) are considered individually as \textit{cold start} approaches.
        \textit{Warm start} approaches, \texttt{Dist-Cross} and \texttt{Cross}, have the number of labeled data points used for assignation; i.e., 1, 3, 5, and 7; as suffixes.
    }
    \label{fig:performance_appendix}
\end{figure*}

\begin{figure*}
    \centering
    \includegraphics[width=1\linewidth]{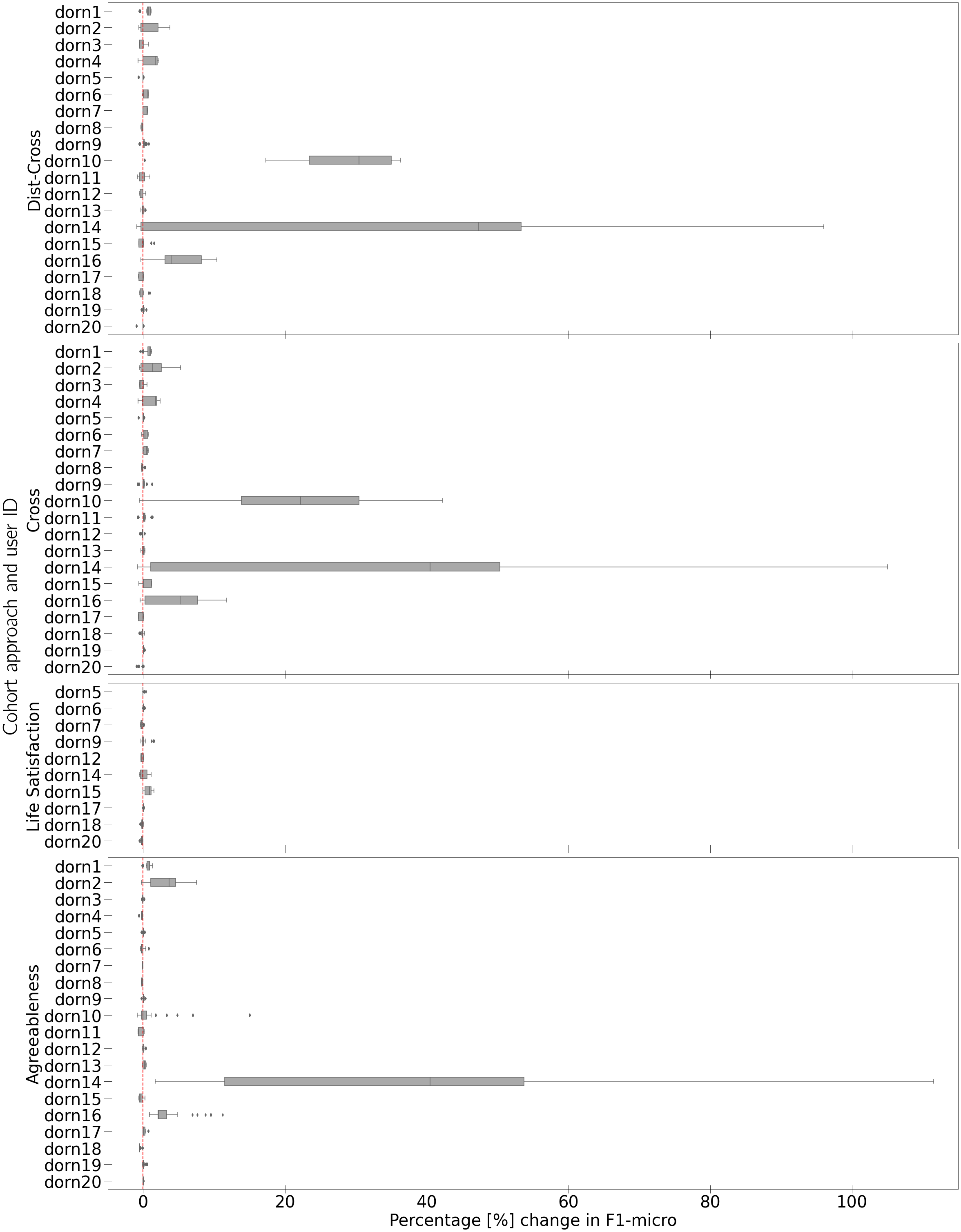}
    \caption{
        Performance percentage change in F1-micro score of 100 iterations for each occupant in each cohort approach (y-axis) for the Dorn dataset.
        The percentage change is calculated based on the respective cohort approach performance and the \texttt{General-purpose} performance.
        The average value of each occupant is reported in Figure~\ref{fig:occupant_change}.
    }
    \label{fig:change_dist_dorn}
\end{figure*}

\begin{figure*}[htb]
    \centering
    \includegraphics[width=1\linewidth]{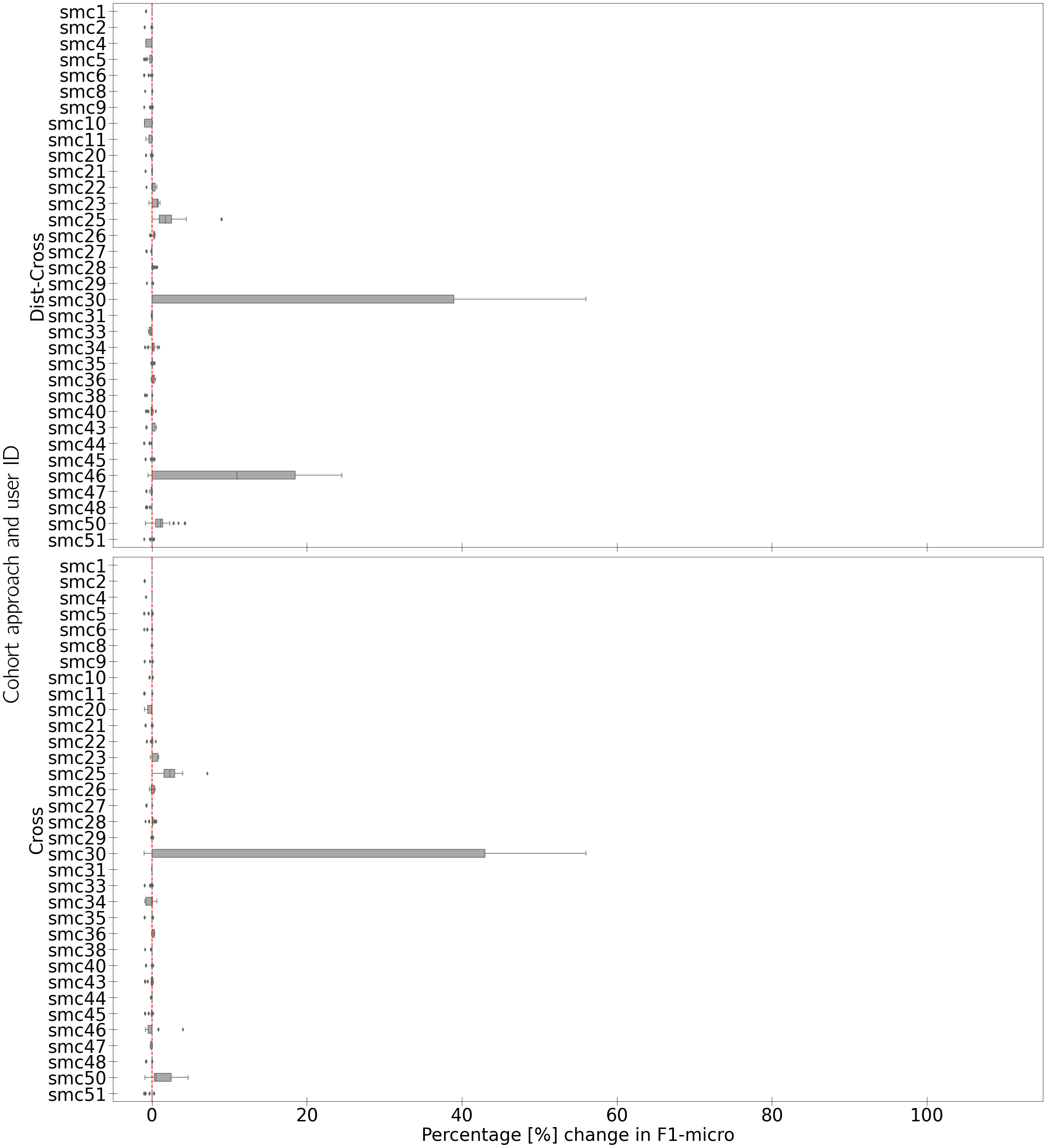}
    \caption{
        Performance percentage change in F1-micro score of 100 iterations for each occupant in each cohort approach (y-axis) for the SMC dataset.
        The percentage change is calculated based on the respective cohort approach performance and the \texttt{General-purpose} performance.
        The average value of each occupant is reported in Figure~\ref{fig:occupant_change}.
    }
    \label{fig:change_dist_smc}
\end{figure*}

\begin{table*}
    \centering
    \begin{tabular}{ccccc}
        \toprule
        Cohort Approach & Better/Worse & Sex & Height (m) & Weight (kgs)\\
        \midrule
        \multirow{2}{*}{Dorn:Dist-Cross} & Better-off & 5M, 4F & 168$\pm$4.71 & 64.3$\pm$12.04\\
        & Worse-off & 5M, 6F & 167$\pm$10.63 & 64.55$\pm$14\\ 
        \midrule
        \multirow{2}{*}{Dorn:Cross} & Better-off & 6M, 6F & 169.08$\pm$8.95 & 64.08$\pm$12.99 \\
        & Worse-off & 4M, 4F & 166$\pm$7.38 & 65$\pm$13.38 \\ 
        \midrule
        \multirow{2}{*}{Dorn:Life Satisfaction} & Better-off & 1M, 3F & 165.25$\pm$8.95 & 55.25$\pm$5.76 \\
        & Worse-off & 3M, 3F & 170$\pm$5.86 & 70.17$\pm$8.53 \\ 
        \midrule
        \multirow{2}{*}{Dorn:Agreeableness} & Better-off & 7M, 4F & 167.18$\pm$8.70 & 65.82$\pm$14.33 \\
        & Worse-off & 3M, 6F & 168.67$\pm$8.16 & 62.78$\pm$11.34 \\
        \midrule
        \multirow{2}{*}{SMC:Dist-Cross} & Better-off & 8M, 4F & 169.58$\pm$11.50 & 74.92$\pm$15.77 \\
        & Worse-off & 8M, 14F & 166$\pm$8.85 & 73.59$\pm$17.02 \\
        \midrule
        \multirow{2}{*}{SMC:Cross} & Better-off & 9M, 2F & 174$\pm$7.36 & 78.36$\pm$18.86 \\
        & Worse-off & 7M, 16F & 164.04$\pm$9.50 & 72$\pm$14.98 \\
        \bottomrule
    \end{tabular}
    \caption{Breakdown of occupants' metadata based on the cohort approaches they are better or worse-off. Height and Weight columns display the mean $\pm$ standard deviation}
    \label{tab:occupants_metadata}
\end{table*}

\clearpage

\bibliographystyle{unsrt}
\bibliography{mybib}

\end{document}